\documentclass[11pt]{article}
\usepackage{acl}
\usepackage{times}
\usepackage{url}
\usepackage{latexsym}
\usepackage{graphicx}
\usepackage{float}
\usepackage{anyfontsize}

\title{Chinese Sentences Similarity via Cross-Attention Based Siamese Network}
\author{Zhen Wang$^1$, Xiangxie Zhang$^2$, Yicong Tan$^3$ \\
Delft University of Technology\\
  { \{$^1$z.wang-42, $^2$x.zhang-60, $^3$y.tan-2\}@student.tudelft.nl}
}
\date{}

\begin{document}

\maketitle

\begin{abstract}

Measuring sentence similarity is a key research area nowadays as it allows machines to better understand human languages. In this paper, we proposed a \textbf{C}ross-\textbf{AT}tention \textbf{S}iamese \textbf{Net}work (CATsNet) to carry out the task of learning the semantic meanings of Chinese sentences and comparing the similarity between two sentences. This novel model is capable of catching non-local features. Additionally, we also tried to apply the long short-term memory (LSTM) network in the model to improve its performance. The experiments were conducted on the LCQMC dataset and the results showed that our model could achieve a higher accuracy than previous work.
\end{abstract}

\section{Introduction}

\begin{figure}[]
    \centering
    \includegraphics[width=0.45\textwidth]{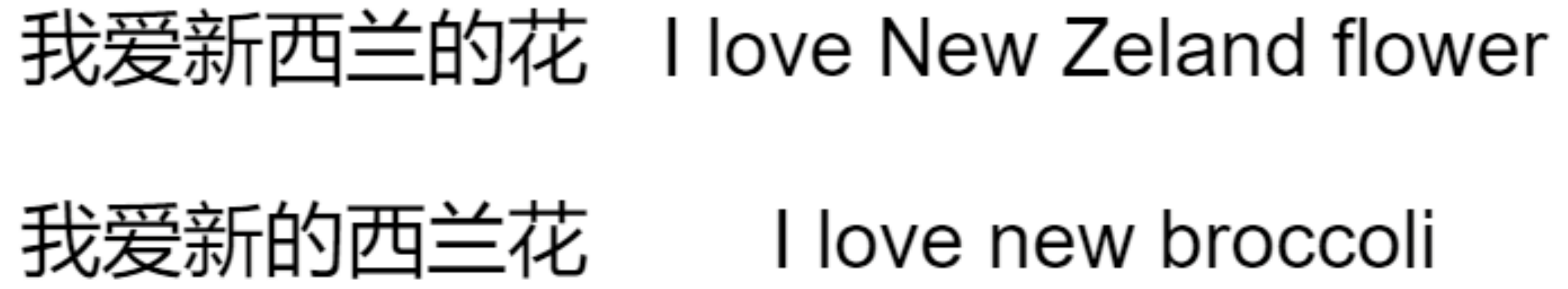}
    \caption{An example of sentences that have similar character form but complete different meanings in Chinese}
    \label{fig:sentence}
\end{figure}

Sentence similarity (STS) is a traditional and still active research direction in Natural Language Processing (NLP). It plays an essential role in many NLP tasks, such as Information Retrieval (IR), Question-Answering (QA), and Dialogue System (DS). The main goal of STS is to determine whether the given two sentences are related or not. Most STS dataset have two relations, relevant or irrelevant. A relevant relation means that the given two sentences talk about the same thing, while irrelevant relation indicates that they express different meanings. Traditional machine learning approaches measure the similarity between sentences based on word-level features. For example, TF-IDF-based features has been applied in sentence similarity measure to detect possible plagiarism \cite{zubarev2014using}. Another example is using semantic distance between words to measure the sentence similarity \cite{miura2015wsl}. However, in most cases, only using the word-level features is not sufficient to capture the meaning of the whole sentence and therefore the similarity measure may not be accurate. Things could be even worse in Chinese language, where the smallest semantic unit is character instead of word as in many European languages. Figure \ref{fig:sentence} shows an example of two Chinese sentence which have extremely similar character form but completely different meanings. However, if the model only takes character-level features into account, then it will probably believe that these two sentences are similar to each other, because they have exactly same characters and only two characters are positioned differently. This example clearly shows the drawback of only using word-level or character-level features, especially when we want to measure the similarity between two sentences. In 2017, the idea of self-attention \cite{vaswani2017attention} mechanism was raised and an architecture named transformer was developed, which models the dependencies between words or characters. It outperformed many previous state-of-the-art algorithms in several NLP tasks. Inspired by transformer, we proposed a cross-attention architecture which specifically aims for modeling the dependencies between the characters across two sentences. We developed a siamese network using the encoder block of the transformer model. Siamese network consists of two exactly the same networks by weight sharing, which has been proved to be effective when the inputs are similar to each other. It was first proposed to verify signatures \cite{bromley1994signature}. We also tried to improve the model by replacing the feed forward network (FFN) in the encoder block of the transformer model with an Bi-LSTM layer.

\begin{figure*}[]
    \centering
    \includegraphics[width=1.0\textwidth]{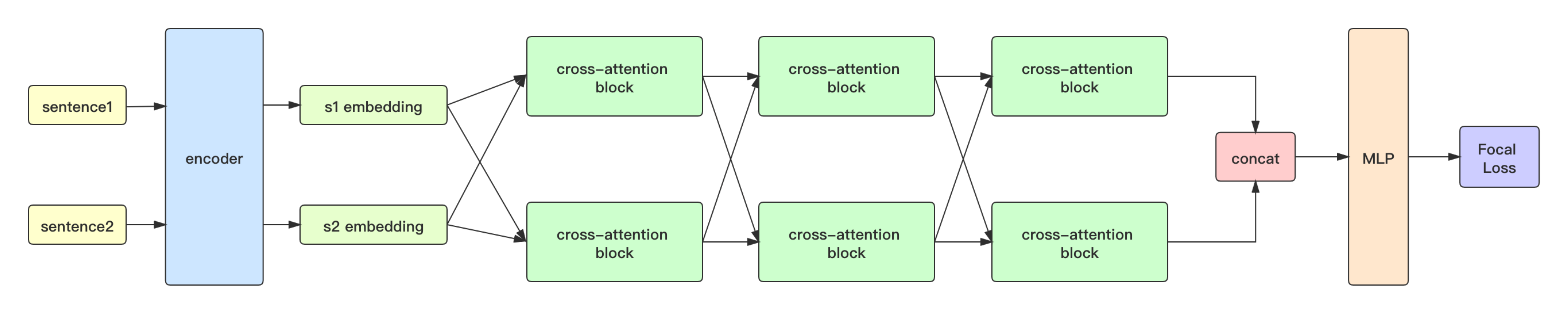}
    \caption{The cross-attention siamese network}
    \label{fig:pipeline}
\end{figure*}

\section{Related work}

Much previous work has focused on applying deep learning models in the STS problem or something similar recently. In 2015, a tree-based convolutional neural network was proposed to capture sentence-level semantic meanings and carry out the task of heuristic matching \cite{mou2015natural}. The time series models, such as RNN, LSTM or GRU, were also applied in this task. Hierarchical Bi-LSTM model with max pooling layer was implemented to find sentence-level representations and it performed well on the sentence similarity task \cite{talman2019sentence}. Similar to our model that uses self-attention mechanism, a dynamic self-attention was proposed for computing sentence embedding, which formed a solid foundation for comparing the similarity between different sentences. There were also plenty of research that concentrate on building siamese models to solve the sentence similarity challenges. In 2016, a siamese recurrent architecture was built for learning sentence similarity \cite{mueller2016siamese}. Following the similar idea, a siamese network which combines CNN and LSTM was implemented to learn semantic textual similarity \cite{pontes2018predicting}. One recent research, which also conducted experiments on the Chinese language, applied siamese CNN architecture for modeling the sentence similarity \cite{shi2020siamese}. Based on these previous work, we built a siamese network where the encoder block of the transformer model was used as the basic model. The next section introduces our model in  detail.

\section{Approach}

The general structure of our model is shown below in Figure \ref{fig:pipeline}, where we proposed the cross-attention block as the fundamental model. 

This was inspired by the self-attention mechanism originally implemented in the model named transformer \cite{vaswani2017attention}. It is an revolutionary model in NLP as it allows the model to encode a sentence globally without using complex recurrent neural network models. This is realized by applying the self-attention mechanism in its encoder. When applying the transformer model, each words needs to be represented by the corresponding word vector. A sentence is thus represented by a two dimensional matrix where each row is a word vector. The input to the encoder block of the transformer model is the 2D matrix. In the self-attention layer, three matrices, namely the query matrix (Q), the key matrix (K), and the value matrix (V) are computed by applying matrix multiplications on the input matrix. The output of the self-attention layer is computed by using the following formula (1) \cite{vaswani2017attention}. In most tasks, single-head attention may not be sufficient to encode the sentence. An improved model is the multi-head attention layer in which several groups of Q, K, V matrices are computed. 

\begin{equation}
    Attention(Q, K, V) = softmax(\frac{QK^T}{\sqrt{d_k}})V
\end{equation}

\begin{figure}[]
    \centering
    \includegraphics[width=0.45\textwidth]{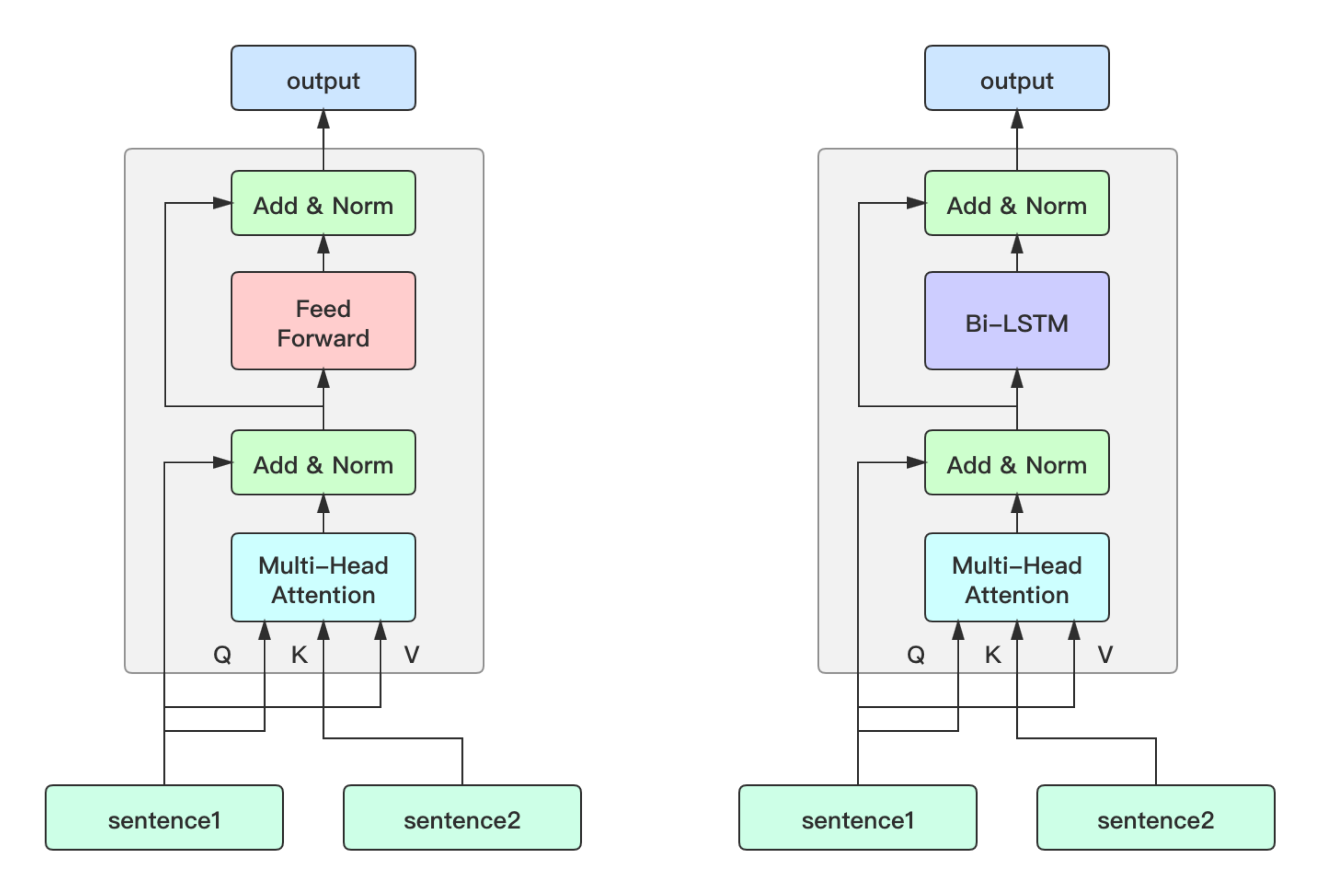}
    \caption{Detail of the Cross-Attention Block. The left one uses MLP and the right one uses Bi-LSTM}
    \label{fig:cblock}
\end{figure}

Our cross-attention block follow the similar idea of the self-attention mechanism. However, unlike the self-attention layer where  the Q, K, V matrix are computed from the same matrix, the cross-attention layer uses two sentence embeddings to calculate these three matrices. Cross-attention was used since we think that the sentence similarity problem requires modeling the relation between the two sentences that need to be compared. Given this challenge, solely applying self-attention is not enough to acquire the dependencies between sentences. The details of the cross-attention block are displayed in the left model in Figure \ref{fig:cblock}. The embedding of one sentence is used to calculate the query and value matrix while the embedding of the other sentence is used to compute the key matrix. In this project, the Baidu Encyclopedia was used for embeddings, where each Chinese character is represented by a 300-D vector. Similar to the self-attention layer in transformer, multi-head attention is used in our model. The rest of the cross-attention block is the as the encoder block of transformer. The idea from the residual network \cite{he2016deep} is used here that the input is added to the output. After the multi-head attention layer is the feed forward layer which is made up with MLP.

\begin{figure}[]
    \centering
    \includegraphics[width=0.45\textwidth]{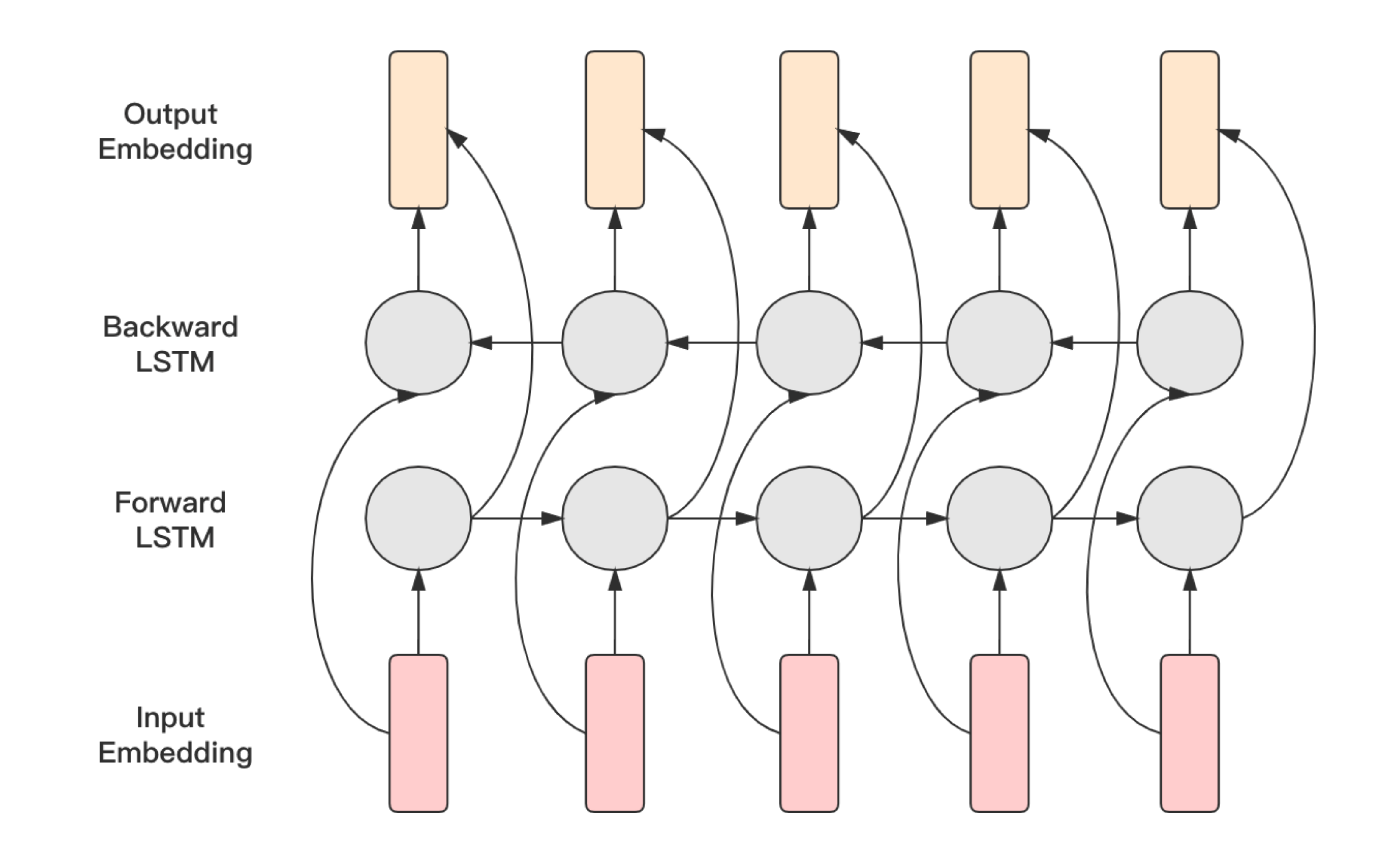}
    \caption{The structure of the bidirectional LSTM layer}
    \label{fig:bilstm}
\end{figure}

To further improve the performance of our model, we tried to replace the MLP layer in the cross-attention block with a bidirectional LSTM layer, as shown in the right side of the above Figure \ref{fig:cblock}. The structure of a Bi-LSTM layer is shown in Figure \ref{fig:bilstm}. There is one forward layer and one backward layer in the Bi-LSTM architecture. By applying the Bi-LSTM layer, we wanted to make our model more robust that can handle more complex dependencies in both directions of the context.

The overall architecture shown in Figure \ref{fig:pipeline} should be clear now. Given two sentences that need to be compared, we first get their sentence embeddings using character vectors. Afterwards, the sentence embeddings are fed into the siamese network which consists of three cross-attention blocks stacking together. One side of the siamese uses the first sentence to compute the query and key matrix and the second sentence to compute the value matrix. However, in the other way around, the other side of the siamese network uses the second sentence for the query and key matrix and the first sentence for calculating the value matrix. By using the siamese network in this way, the dependencies between the two sentences in both directions can be modeled. For each pair of sentences, there are two outputs from the siamese network. The two outputs are concatenated and then passed through an MLP layer to get the final result. We used the focal loss \cite{lin2017focal} as our loss function. It is proved to be helpful in alleviating the impact caused imbalanced data. The formula of the focal loss function is shown below in formula (2). It is an updated version of the cross entropy loss where there are tunable parameters. 
\begin{equation}
    FL(p_t) = -\alpha_t(1-p_t)^\gamma log(p_t)
\end{equation}

\section{Experiments and Results}

\subsection{Datasets and Metrics}

The lack of dataset for non-English language has been a great challenge for NLP researchers. In 2018, a large-scale Chinese question matching corpus (LCQMC) was built \cite{liu2018lcqmc}. Each data sample consists of a pair of questions and a number indicates whether these two questions are similar or not. 0 means the two questions are not similar while 1 means they are similar. The questions were collected from questions answering website in Chinese, for example Baidu Knows and Zhihu, which are similar to Quora. There are numerous questions on this kind of website and many of them are duplicated and have highly similar meanings. Therefore, this dataset is very suitable for the sentence similarity task. We used this dataset to train our model. 238766 question pairs were used for training while 12500 question pairs were used for testing the model. The accuracy is used as the evaluation metric for the model. 

\subsection{Quantitative Results}

In total, we built three models for comparisons. The first model is a self-attention siamese network where one sentence is used for computing the query, key and value matrix. The next two models have been discussed in the previous section, who use three cross-attention blocks. One uses MLP after the cross-attention layer while the other one uses Bi-LSTM. We compared our models with three baseline models on the sentence similarity task. Those are siamese LSTM architecture \cite{mueller2016siamese}, siamese Bi-LSTM architecture \cite{ranasinghe2019semantic}, and siamese convolutional architecture \cite{shi2020siamese}. The results are shown in Table \ref{tab:results}. The top half shows the results of the baseline models while the bottom half displays the results of our models.

\begin{table}[!htb]
\centering
\begin{tabular}{cc}
\hline
Architecture & Accuracy \\ \hline
Siamese LSTM  & 68.63 \\
Siamese Bi-LSTM   & 68.64 \\
Siamese Convolutional  & 77.31 \\ \hline
Siamese Self-Attention  & 78.61 \\
Siamese Cross-Attention MLP  & 81.99 \\
CATsNet (Ours)  &  \textbf{83.15}\\ \hline
\end{tabular}
\caption{The results on the LCQMC dataset}
\label{tab:results}
\end{table}

\subsection{Ablation Analysis}

\begin{figure}[!htb]
    \centering
    \includegraphics[width=0.5\textwidth]{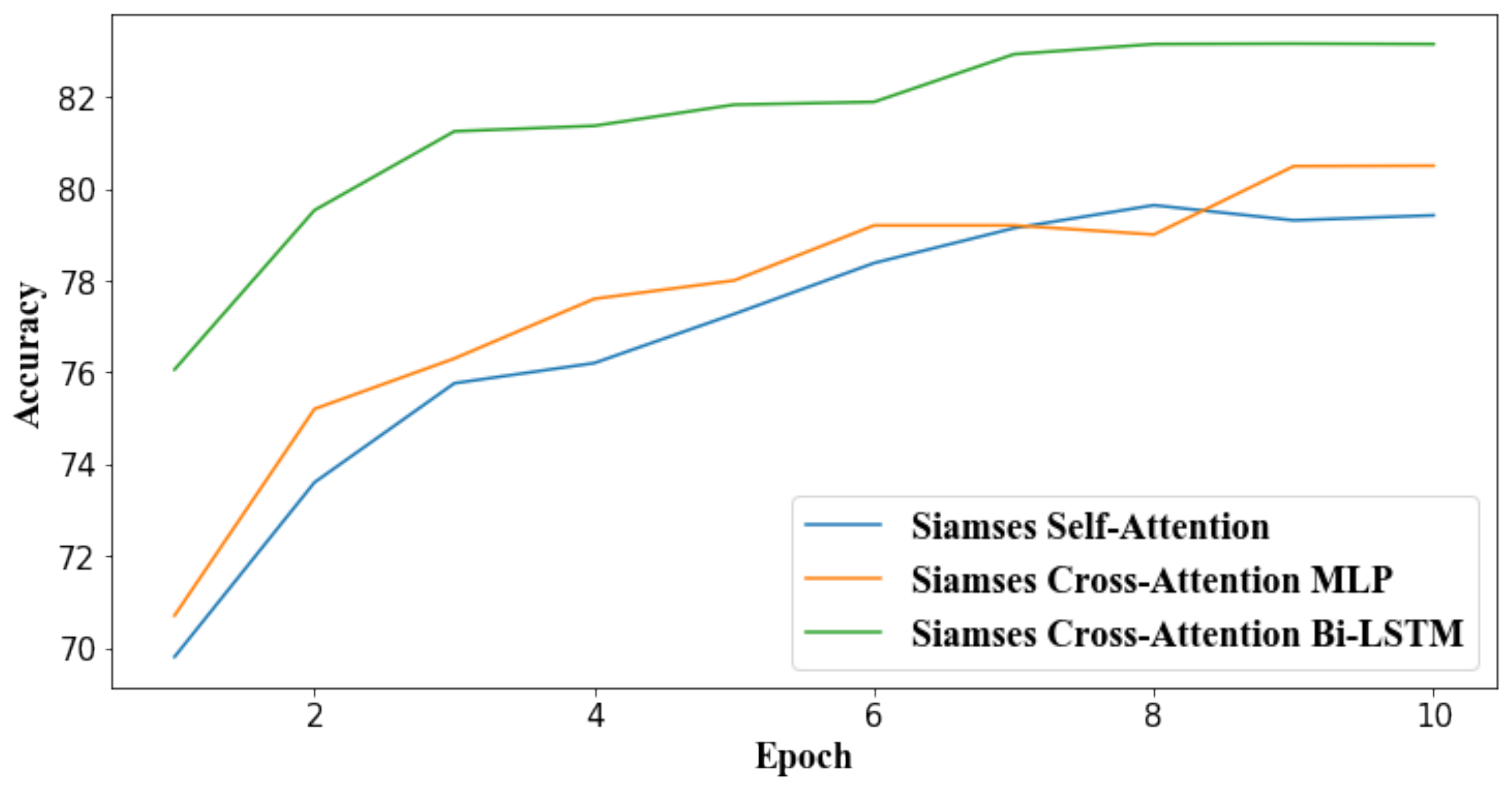}
    \caption{The accuracy on validation dataset}
    \label{fig:acc}
\end{figure}

We firstly compared the Siamese Self-Attention model with the previous siamese models, including the LSTM-based, the Bi-LSTM-based and the CNN-based model. The final result shows that Self-Attention can achieve an accuracy of 78.61\%, which outperforms previous models. This suggests that in the STS task, self-attention mechanism is more useful than RNN-based or CNN-based model. We argue that this is because that the self-attention mechanism can capture the global dependencies between words in a sentence better than other models. We then compared our cross-attention model with the self-attention model. The result illustrates that the cross-attention model can generate better result with an accuracy of 81.99\%. This provides a solid evidence on the effectiveness of our proposed cross-attention model. The reason behind this is that cross-attention not only models the dependencies between words in the same sentence, but also the dependencies between words across the two sentences. Given that sentence similarity challenge is related to two sentences, our cross-attention architecture encodes more information than the self-attention architecture. In the last experiment, we used the CATsNet, which replaces the MLP layer in cross-attention block with a Bi-LSTM layer. The result shows that CATsNet perform the best among all models with an accuracy of 83.15\%. We argue that this is because, while the cross-attention mechanism can make use of non-local features in the sentences, Bi-LSTM module can help us to extract sentences' sequential features. Combining the global features and sequential features between and within sentences is supposed to give us the best representations of sentences. To better understand the effectiveness of our model, we in addition draw the accuracy curve in validation dataset, which is shown in Figure \ref{fig:acc}. This figure illustrates that being trained with same number of epochs, CATsNet can always achieve better performance than other siamese models. In conclusion, our proposed model is robust and can work well on the sentence similarity task on the LCQMC dataset. 

\section{Conclusions}
In this paper, we proposed the CATsNet architecture which is a siamese network. It uses the cross-attention layer and the Bi-LSTM layer to better encode the information of two sentences. The results show that our CATsNet can outperform other siamses-based models. The use of focal loss also helps to improve the performance. This project provides convincing evidence on the effectiveness of applying siamese models on NLP tasks. In the future, it would be interesting to apply this proposed model on more languages such as English. By doing such, we could check the generality of our model.






\bibliographystyle{acl}
\bibliography{acl}

\end{document}